%% file: Extended-LaTeX-2026.tex
\documentclass[letterpaper]{article} 
\usepackage{aaai2026}  
\usepackage{times}  
\usepackage{helvet}  
\usepackage{courier}  
\usepackage[hyphens]{url}  
\usepackage{graphicx} 
\usepackage{natbib}  
\usepackage{caption} 
\usepackage{amsmath}
\usepackage{booktabs} 
\usepackage{multirow}
\usepackage{amssymb}
\usepackage{fontawesome}
\usepackage{makecell}
\usepackage{adjustbox}
\usepackage{graphicx}
\usepackage{caption}
\usepackage{subfig}   
\usepackage{xcolor}      
\usepackage{array}    
\usepackage{colortbl}  
\usepackage{arydshln}
\usepackage{algorithm2e}

\usepackage{algorithm}
\usepackage{algorithmic}

\usepackage{newfloat}
\usepackage{listings}
\DeclareCaptionStyle{ruled}{labelfont=normalfont,labelsep=colon,strut=off} 
\floatstyle{ruled}
\newfloat{listing}{tb}{lst}{}
\floatname{listing}{Listing}
\title{Point Cloud Quantization through Multimodal \\ Prompting for 3D Understanding}
\author{
    Hongxuan Li\textsuperscript{\rm 1},
    Wencheng Zhu\textsuperscript{\rm 1,2}\thanks{Corresponding author},
    Huiying Xu\textsuperscript{\rm 3},
    Xinzhong Zhu\textsuperscript{\rm 3},
    Pengfei Zhu\textsuperscript{\rm 1}
}
\affiliations{
    \textsuperscript{\rm 1}College of Intelligence and Computing, Tianjin University\\
    \textsuperscript{\rm 2}Haihe Laboratory of Information Technology Application Innovation\\
    \textsuperscript{\rm 3}School of Computer Science and Technology, Zhejiang Normal University\\
    \{lihongxuan, wenchengzhu, zhupengfei\}@tju.edu.cn,
    \{xhy, zxz\}@zjnu.edu.cn
}

\usepackage{bibentry}

\begin{document}

\maketitle

\begin{abstract}
Vector quantization has emerged as a powerful tool in large-scale multimodal models, unifying heterogeneous representations through discrete token encoding.
However, its effectiveness hinges on robust codebook design.
Current prototype-based approaches relying on trainable vectors or clustered centroids fall short in representativeness and interpretability, even as multimodal alignment demonstrates its promise in vision-language models.
To address these limitations, we propose a simple multimodal prompting-driven quantization framework for point cloud analysis. 
Our methodology is built upon two core insights: 1) Text embeddings from pre-trained models inherently encode visual semantics through many-to-one contrastive alignment, naturally serving as robust prototype priors; and 2) Multimodal prompts enable adaptive refinement of these prototypes, effectively mitigating vision-language semantic gaps.
The framework introduces a dual-constrained quantization space, enforced by compactness and separation regularization, which seamlessly integrates visual and prototype features, resulting in hybrid representations that jointly encode geometric and semantic information.
Furthermore, we employ Gumbel-Softmax relaxation to achieve differentiable discretization while maintaining quantization sparsity.
Extensive experiments on the ModelNet40 and ScanObjectNN datasets clearly demonstrate the superior effectiveness of the proposed method.
\end{abstract}

\begin{links}
    \link{Code}{https://github.com/li-hongxuan/PCQ}
\end{links}

\section{Introduction}
Human cognition organizes concepts not by rigid definitions, but through prototypes, where their semantic meaning emerges from graded similarity to exemplary instances \cite{hampton2006concepts}. Linguistic studies characterize prototypes by \textit{vagueness} (ambiguous boundaries), \textit{typicality} (graded membership), \textit{genericity} (class-wide applicability), and \textit{opacity} (non-transparent categorization) \cite{geeraerts2006prototype}. Interestingly, textual descriptions inherently exhibit these prototype characteristics. Language shapes concepts through hierarchical abstraction while maintaining fuzzy boundaries \cite{xu2025pointllm}. This inherent semantic isomorphism raises a critical question for multimodal learning: \textit{Given that text embeddings possess an intrinsically prototype-like structure, can they serve as a bridge between visual perception and conceptual understanding?} 

\begin{figure}[!t]
\centering
\includegraphics[width=1\columnwidth]{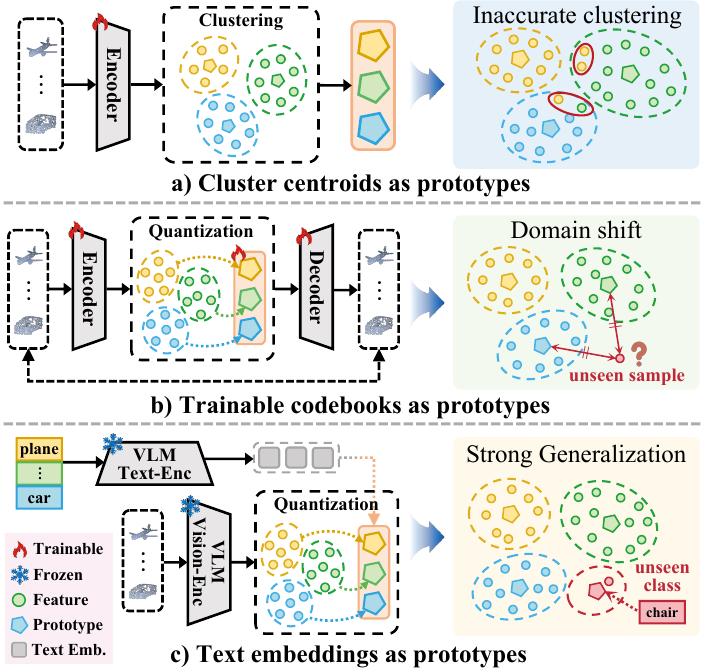} 
\caption{
a) Cluster centroids as prototypes and b) Trainable codebooks as prototypes suffer from inaccurate clustering and domain shift, which reduces their representativeness and generalization. c) Our method leverages a pre-trained vision-language model to derive text-driven semantic prototypes, refined during fine-tuning to enhance representativeness, interpretability, and generalization for 3D understanding.
}
\label{fig1}
\end{figure}

Recent advancements in large multimodal models, such as CLIP \cite{radford2021learning}, ULIP \cite{xue2023ulip}, and Uni3D \cite{zhou2023uni3d} have revolutionized 2D/3D understanding \cite{cho2023distribution} by aligning heterogeneous modalities into a unified embedding space through contrastive learning \cite{li2022blip, zeng2023clip2}. These models achieve notable performance across various tasks, from image classification \cite{chen2023clip2scene} to point cloud understanding \cite{zhang2024towards}. 
Despite such progress, a fundamental misalignment is inherent: while visual encoders excel in feature extraction \cite{ning2024dilf, guo2023point, srivastava2024omnivec}, text encoders primarily focus on modeling high-level semantic hierarchies \cite{gan2022vision, palanisamy2023proto}. 
This limited alignment capability of multimodal models often results in degraded performance in downstream tasks.

To address semantic gaps in cross-modal understanding, recent research has turned to prototype learning \cite{wei2023online,yang2022visual}. 
Existing methods fall into cluster-based and codebook-based categories.
Cluster-based approaches employ cluster centroids derived from training data as representative prototypes \cite{li2021prototypical}. While intuitive, such centroids are constrained by data distribution and initialization, failing to capture intra-class diversity and thus limiting representativeness and generalization \cite{shu2022test}.
In contrast, codebook-based methods learn a set of prototypes by optimizing task-specific objectives \cite{van2017neural}.
Although flexible, these methods often suffer from domain shift and unstable convergence, while offering limited interpretability.
Crucially, neither category effectively leverages cross-modal knowledge to enable robust and generalizable prototype learning.

In this paper, we introduce a method to effectively utilize text-guided prototypes derived from pre-trained vision-language models. 
Our approach stems from two key observations: 
1) Vision-language models achieve point cloud-text alignment through many-to-one contrastive learning, where diverse 3D instances of the same class are aligned with a single text embedding, e.g., "a 3D shape of a chair". This directly mirrors prototype characteristics of vagueness (tolerance for intra-class variance) and genericity (class-wide applicability). 
2) Text embeddings inherently possess prototype-like properties such as typicality (graded similarity to class exemplars) and opacity (implicit categorization), making them well-suited as semantic prototypes for visual representation learning.
By reformulating text embeddings as trainable prototypes, our framework enables dynamic integration of visual features with language-derived conceptual knowledge, effectively bridging the vision-language domain gap.
To further enhance vision-language alignment, we refine these prototypes using learnable multimodal prompts, optimized under compactness and separation constraints. 
We then quantize point cloud features into discrete prototypes via Gumbel-Softmax to generate text-like features. 
Finally, a cross-modal feature fusion module combines fine-grained visual details with high-level semantic concepts.

Our main contributions are summarized as follows:
\begin{itemize}
    \item We propose a text-driven 3D quantization framework that unifies vision-language alignment through point cloud feature discretization.

    \item We reformulate text features as trainable prototypes and model their quantization via Gumbel distribution for end-to-end gradient estimation, while preserving discrete semantics for cross-modal generalization. 

    \item Extensive experiments on the ModelNet40 and ScanObjectNN datasets demonstrate superior performance compared to state-of-the-art methods while maintaining competitive parameter efficiency.   
    
\end{itemize}

\section{Related Work}
\noindent
\textbf{3D Multi-Modality Models.}
Extending large multimodal models to point cloud analysis presents significant challenges \cite{qian2022pointnext, zhou2023partslip++}. Recent approaches can be broadly categorized into two paradigms: projection-based \cite{ma2022rethinking} and unified representation learning methods \cite{huang2024joint,qi2024gpt4point}.
Projection-based approaches, such as PointCLIP \cite{zhang2022pointclip} and PointCLIP V2 \cite{zhu2023pointclip}, transform 3D point clouds into multi-view 2D images and leverage pre-trained vision-language models.
CLIP2Point \cite{huang2023clip2point} extends this pipeline by fusing depth and RGB features across views. 
Duoduo CLIP \cite{lee2025duoduo} incorporates multi-view projections with spatial context.
However, these approaches inherently suffer from information loss during the 2D projection process \cite{hegde2023clip,hess2024lidarclip}.
In contrast, ULIP \cite{xue2023ulip, xue2024ulip}, Openshape \cite{liu2023openshape}, and Uni3D \cite{zhou2023uni3d} achieve cross-modal alignment by training 3D encoders on triplets of point clouds, images, and text.
PPT \cite{sun204parameter} employs parameter-efficient tuning of pre-trained 3D encoders. 
Despite effectiveness, these methods face a semantic gap that visual encoders prioritize object perception \cite{huang2024frozen, zhang2023clip}, while text encoders capture abstract concepts \cite{xu2023multimodal}.
Recent work explores prototype-based methods. ProtoCLIP \cite{chen2023protoclip} learns visual prototypes through contrastive language guidance, while Partslip \cite{liu2023partslip} aligns part-level 3D features with textual descriptions.

\noindent
\textbf{Vector Quantization.}
Vector quantization is a prominent topic in representation learning.
Existing VQ methods can be categorized into deterministic quantization and stochastic quantization. 
Deterministic quantization \cite{zhan2022auto}, such as VQ-VAE \cite{van2017neural}, selects codebook entries via argmax/min operations and has seen extensions like multiscale codebooks in VQ-VAE2 \cite{razavi2019generating}, improved training stability in DVAE \cite{vahdat2018dvae}, and integration of adversarial losses in VQ-GAN \cite{esser2021taming}. VIM \cite{yu2021vector} optimizes codebook learning efficiency and reconstruction fidelity.
Stochastic quantization samples tokens from learned probability distributions \cite{maddison2014sampling}, thus requiring gradient estimation techniques for non-differentiable problems.
VQ-Wave2Vec \cite{baevski2019vq} employs Gumbel-Softmax reparameterization \cite{jang2016categorical} for differentiable sampling.
DALL-E \cite{ramesh2021zero} leverages stochastic quantization for text-to-image generation. 
Theoretical grounding for soft assignment has also been provided \cite{roy2018theory}. 
However, deterministic approaches often suffer from codebook collapse and limited semantic expressiveness, while stochastic variants can struggle with training instability  \cite{zhang2024conceptual}. 
Crucially, neither category learns codebooks that sufficiently represent the rich semantics of language.
To address these limitations, our framework redefines prototypes through text-guided quantization.

\begin{figure*}[t] 
\centering 
\includegraphics[width=0.97\linewidth]{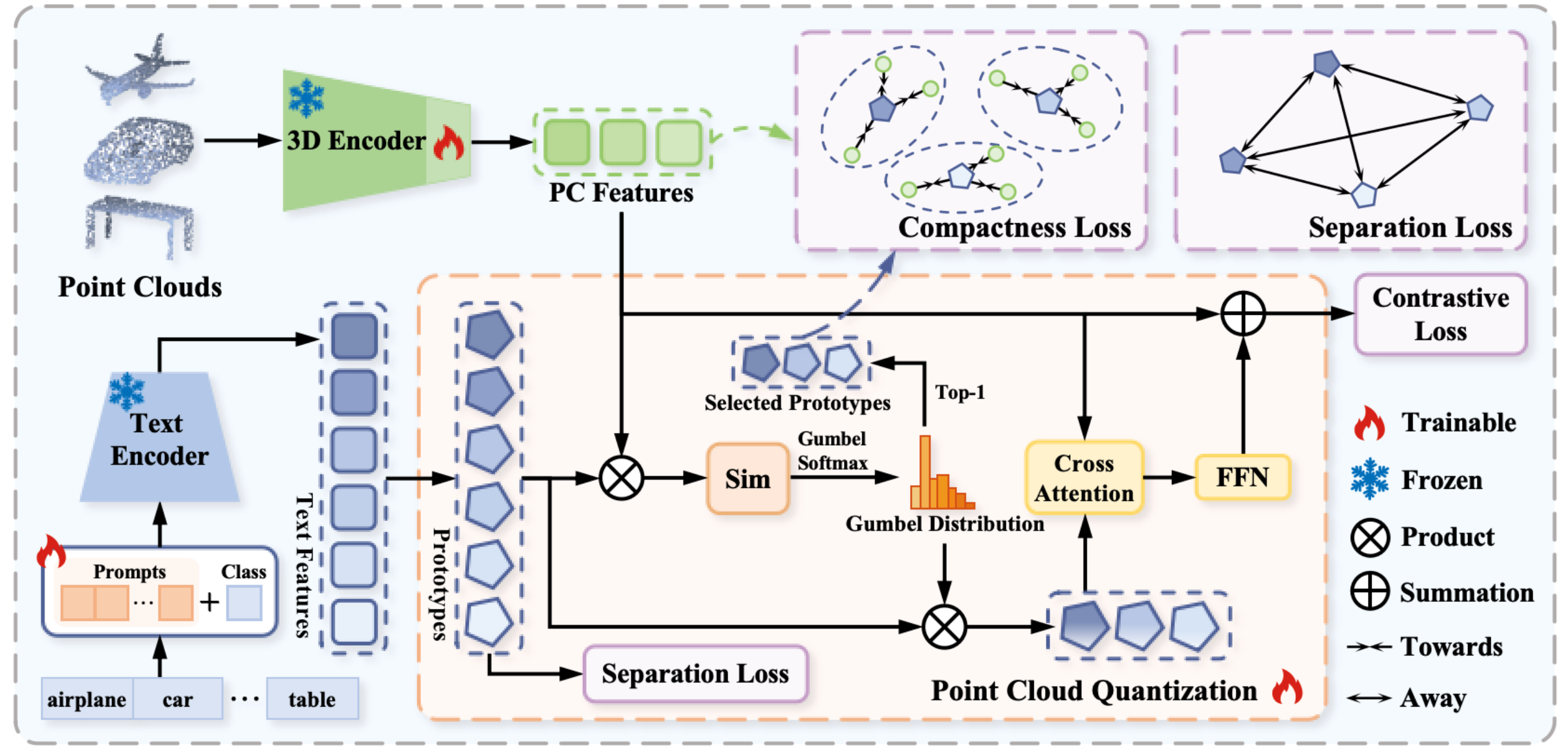} 
\caption{
Framework of the proposed approach. Our method comprises feature extraction and point cloud quantization modules. The feature extraction module uses ULIP-2 text encoder and 3D point cloud encoder to extract text and point cloud features. The quantization module then takes these text features as prototypes and quantizes point cloud features into prototype features. To enable differentiable sampling, discrete features are modeled through a Gumbel distribution, and Gumbel-Softmax reparameterization is adopted to represent point cloud features with prototype features. Finally, point cloud features are combined with prototype features via cross-modal feature fusion to produce the final hybrid representation.
Notably, parameter-efficient fine-tuning is employed to optimize both prototype and point cloud features, constrained by compactness and separation losses.
} 
\label{framework} 
\end{figure*}

\section{Methodology}

As illustrated in Figure \ref{framework}, our framework introduces a dual-stream architecture that connects visual perception and semantic abstraction through a text-guided prototype learning paradigm.
The key innovation lies in redefining text embeddings from pre-trained vision-language models as trainable visual prototypes, leveraging the semantic hierarchy learned through contrastive alignment pretraining.

Our Point Cloud Quantization (PCQ) framework converts continuous features into discrete semantic tokens.
In our experiments, we adopt ULIP-2 as the pre-trained multimodal backbone because it achieves comparable performance to Uni3D-Ti while using fewer parameters, making it suitable for memory-limited scenarios.
To enable adaptive prompt tuning, we introduce learnable prompts optimized with two complementary regularizers: compactness loss to encourage intra-prototype coherence and separation loss to promote inter-prototype diversity. 
Together, these objectives drive the prototypes toward representations that reduce modality misalignment.
Next, we take the textual embeddings as visual prototypes and quantize the visual features into discrete tokens via Gumbel-Softmax reparameterization. 
Finally, we fuse original point-cloud features with their corresponding prototypes to produce the downstream task representation. 

\subsection{Preliminaries}

ULIP-2 constructs a large-scale dataset of aligned text-image-point cloud triplets derived from Objaverse \cite{deitke2023objaverse} to facilitate joint pre-training of a unified representation.
Each triplet $U_i=(I_i, T_i, P_i)$ comprises multi-view rendered images $I_i$ of the 3D object, textual descriptions $T_i$ of images, and sampled point cloud $P_i$. ULIP-2 employs three modality-specific encoders:
a vision encoder $\mathcal{F_I}(\cdot)$ and a text encoder $\mathcal{F_T}(\cdot)$, both initialized from SLIP \cite{mu2022slip}, and a point cloud encoder $\mathcal{F_P}$ based on the trainable PointBERT \cite{yu2022point} backbone.
The weights of $\mathcal{F_I}$ and $\mathcal{F_T}$ are initialized and kept frozen to preserve semantic alignment, while the 3D encoder $\mathcal{F_P}$ is trainable.
The embedding $U_i$ is formalized as,
\begin{equation}
\textbf{h}^I_i, \textbf{h}^T_i, \textbf{h}^P_i = \mathcal{F_I}(I_i), \mathcal{F_T}(T_i), \mathcal{F_P}(P_i).
\end{equation}
During pretraining, ULIP-2 optimizes cross-modal contrastive losses between modalities $m_1$ and $m_2$ as, 
\begin{equation}
\begin{split}
\mathcal{L}_{m_1, m_2}  =  - \frac{1}{2}  \sum_{(i,j)} & \left[  \log \frac{\exp(f(\textbf{h}^{m_1}_i, \textbf{h}^{m_2}_j))}{\sum_k \exp(f(\textbf{h}^{m_1}_i, \textbf{h}^{m_2}_k))} \right.\\ 
  &  \left. +  \log \frac{\exp(f(\textbf{h}^{m_1}_i, \textbf{h}^{m_2}_j))}{\sum_k \exp(f(\textbf{h}^{m_1}_k, \textbf{h}^{m_2}_j))} \right ],
\end{split}
\end{equation}
where $(i,j)$ represents positive pairs in each training batch, and $f(\textbf{a},\textbf{b})=\textbf{a}^\top\textbf{b}$ computes cosine similarity between two input features. 
The final loss combines all modality pairs through weighted summation,
\begin{equation}
\mathcal{L}_{\text{Final}} = \alpha \mathcal{L}_{(I, T)} + \beta \mathcal{L}_{(I, P)} + \gamma \mathcal{L}_{(P, T)},
\end{equation}
with $\alpha$, $\beta$, and $\gamma \in [0,1]$ controlling the relative importance of image-text, image-point, and point-text alignment.


\subsection{Point Cloud Quantization}

\textbf{Adaptive Prompt Tuning.} 
Our framework leverages a pre-trained text encoder $\mathcal{F_T}$ and a 3D point cloud encoder $\mathcal{F_P}$ from ULIP-2 for feature extraction.
To preserve semantic knowledge from large-scale pretraining while addressing vision-language semantic misalignment in downstream datasets, we freeze the text encoder and introduce $m$ learnable prompt tokens for parameter-efficient adaptation. 
The textual prototype $\textbf{h}_k^T $ for the $k$-th class is then computed as, 
\begin{equation}
\textbf{h}_k^T=\mathcal{F}_\mathcal{T}(\textbf{T}_k),~~\mathrm{where } \; \textbf{T}_k=[\textbf{u}_1,\textbf{u}_2,\ldots,\textbf{u}_m,\textbf{c}_k].
\end{equation}
$\textbf{T}_k$ denotes the $k$-th input to the text encoder containing $m$ learnable prompt  vectors $\textbf{u}_1,\textbf{u}_2,\ldots,\textbf{u}_m$, where $k \in [1, K]$ and $K$ is the total number of classes.
Here, $\textbf{c}_k$ represents the text token of the $k$-th class name, e.g., "plane" or "car".
By augmenting the fixed class embedding with learnable prompts, this design enriches the semantic representation with adaptable knowledge, facilitating more effective and flexible prototype generation.

To maintain the generalization capabilities of the pre-trained 3D encoder $\mathcal{F_P}$, we freeze all its layers except the final Transformer block, which is fine-tuned to adapt to task-specific features. The feature extraction is formalized as,
\begin{equation}
\textbf{h}^P_i = \mathcal{F_{P'}}(P_i),~~i \in [1, N],
\end{equation}
where $ \mathcal{F_{P'}}(\cdot) $ denotes the partially fine-tuned 3D encoder, with only the final Transformer block for training. $P_i$ is the $i$-th point cloud instance and $N$ is total number of instances.
This parameter-efficient fine-tuning strategy optimizes the alignment between text and point cloud features.

\noindent
\textbf{Prototype-Guided Differentiable Quantization.} 
A fundamental challenge in multimodal alignment lies in the discrete-continuous gap: text encodes structured semantics through discrete and interpretable tokens, while visual features are inherently continuous, enabling generalization at the cost of inter-class ambiguity. 
To address this gap, we propose using text-derived embeddings as semantic prototypes and quantizing continuous visual features into this textual prototype space. This strategy enhances interpretability by grounding visual representations in human-readable semantic structures while reducing inter-class feature overlap through well-defined and discrete decision boundaries.
However, hard quantization is non-differentiable, hindering end-to-end training. We address this using a differentiable Gumbel-Softmax relaxation that enables soft and probabilistic assignments during training.

For each point cloud instance $P_i$, we compute the pairwise cosine similarity between its feature $\textbf{h}^P_i$ and all text-derived prototypes $\textbf{h}^T_k$,
$s_{ik}$ represents the semantic affinity between $P_i$ and the $k$-th class prototype guiding the quantization of point cloud features into prototype-aligned representations. 
We then model the assignment process using a Gumbel distribution and perform differentiable optimization via the Gumbel-Softmax reparameterization technique \cite{xu2022groupvit}. Formally, the prototype assignment probability is computed as $
q_{ik} = \frac{\exp(s_{ik})}{\sum_{j=1}^K \exp(s_{ij})}$, 
where $q_{ik}$ represents the probability of the $k$-th prototype being assigned to the $i$-th point cloud feature.
Gumbel-Softmax injects stochasticity via additive Gumbel noise as,
\begin{equation}
y_{ik} = \frac{\exp\left(\frac{\log q_{ik} - \log(-\log \epsilon_k)}{\tau}\right)}{\sum_{j=1}^K \exp\left(\frac{\log q_{ij} - \log(-\log \epsilon_j)}{\tau}\right)}.
\end{equation}
$\epsilon_k$ represents noise sampled from a uniform distribution $U[0, 1]$, $\tau$ is a temperature parameter that controls the sharpness of the distribution. 
In our experiments, we adopt a default value of $\tau=1$. 
The quantized probability weight $y_{ik}$ approximates a one-hot prototype selection while retaining differentiability.
The quantized feature $\textbf{v}_i$ is derived as, 
\begin{equation}
\textbf{v}_i=\sum_{k=1}^Ky_{ik}\textbf{h}_k^T.
\end{equation}
  
\noindent
\textbf{Cross-Modal Feature Fusion.}
We fuse the original point cloud feature $\textbf{h}^P_i$ with its quantized feature $\textbf{v}_i$ to integrate geometric details from 3D data with high-level semantic abstractions from language. The resulting hybrid feature $\textbf{f}_i$ preserves spatial structure while being enriched with semantic guidance as
$\textbf{f}_i = \text{FFN}\left(\text{CrossAttention}\left(\textbf{h}^P_i, \textbf{v}_i\right) \right)+\textbf{h}^P_i$.
The \textit{CrossAttention} layer uses $\textbf{h}^P_i$ as queries, and $\textbf{v}_i$ as keys/values to attend to semantically relevant prototypes. A residual connection ensures that geometric information is preserved, while selectively enhancing it with language semantics.

\begin{table*}[h]
    \centering
    \normalsize
    \setlength{\tabcolsep}{4.5pt}  
    \renewcommand{\arraystretch}{1.09}  
    \begin{tabular}{l|ccccc|cccccccc|cc}
        \noalign{\global\setlength{\arrayrulewidth}{0.8pt}}
        \cline{0-15}
        \rule{-2.5pt}{2.5ex}
        & \multicolumn{5}{c|}{\textbf{Supervised Training}} & \multicolumn{8}{c|}{\textbf{Unsupervised Pre-Training + Full Fine-Tuning}} & \multicolumn{2}{c}{\textbf{PEFT}} \\
        \noalign{\global\setlength{\arrayrulewidth}{0.5pt}}
        \cline{2-16}
        \textbf{Dataset}  
        & \makecell[c]{\adjustbox{angle=90,valign=m}{\shortstack{PointNet\\\scriptsize \cite{qi2017pointnet}}}} 
        & \makecell[c]{\adjustbox{angle=90,valign=m}{\shortstack{PointNet++\\\scriptsize \cite{qi2017pointnet++}}}} 
        & \makecell[c]{\adjustbox{angle=90,valign=m}{\shortstack{PointCNN\\\scriptsize \cite{li2018pointcnn}}}}   
        & \makecell[c]{\adjustbox{angle=90,valign=m}{\shortstack{DGCNN\\\scriptsize \cite{wang2019dynamic}}}} 
        & \makecell[c]{\adjustbox{angle=90,valign=m}{\shortstack{MVTN\\\scriptsize (Hamdi et al. 2021)}}} 
        & \makecell[c]{\adjustbox{angle=90,valign=m}{\shortstack{PointBERT\\\scriptsize \cite{yu2022point}}}} 
        & \makecell[c]{\adjustbox{angle=90,valign=m}{\shortstack{MaskPoint\\\scriptsize (Liu et al. 2022)}}} 
        & \makecell[c]{\adjustbox{angle=90,valign=m}{\shortstack{PointMAE\\\scriptsize \cite{pang2022masked}}}} 
        & \makecell[c]{\adjustbox{angle=90,valign=m}{\shortstack{PointCMT\\\scriptsize \cite{yan2022let}}}} 
        & \makecell[c]{\adjustbox{angle=90,valign=m}{\shortstack{PointM2AE\\\scriptsize ~\cite{zhang2022point}~~}}} 
        & \makecell[c]{\adjustbox{angle=90,valign=m}{\shortstack{ACT\\\scriptsize \cite{dong2022autoencoders}}}} 
        & \makecell[c]{\adjustbox{angle=90,valign=m}{\shortstack{ULIP\\\scriptsize \cite{xue2023ulip}}}} 
        & \makecell[c]{\adjustbox{angle=90,valign=m}{\shortstack{ULIP-2\\\scriptsize \cite{xue2024ulip}}}} 
        & \makecell[c]{\adjustbox{angle=90,valign=m}{\shortstack{PPT*\\\scriptsize \cite{sun204parameter}}}} 
        & \makecell[c]{\adjustbox{angle=90,valign=m}{\shortstack{\textbf{PCQ}\\(Ours)}}} \\
        \noalign{\global\setlength{\arrayrulewidth}{0.5pt}}
        \cline{0-15}
        \rule{-2.5pt}{2.5ex}
        \textbf{MN40}~\cite{wu20153d}& 89.2 & 90.7 & 92.2 & 92.9 & 93.5 & 93.2 & 93.8 & 93.8 & 93.5 & 93.4 & 93.7 & \underline{94.1} & -- & 93.6 & \textbf{94.1} \\
        \textbf{OBJ}~\cite{uy2019revisiting} & 79.2 & 84.3 & 85.5  & 86.2 & 92.3 & 88.1 & 89.7 & 88.3 & 92.3 & 88.8 & 91.9 & -- & -- & \underline{93.1} & \textbf{93.5}  \\
        \textbf{BG}~\cite{uy2019revisiting} & 73.3 & 82.3 & 86.1 & 82.8 & 92.6 & 87.4 & 89.3 & 90.0 & 92.6 & 91.2 & 93.3 & -- & -- & \underline{95.4} & \textbf{95.5} \\
        \textbf{PB}~\cite{uy2019revisiting} & 68.0 & 77.9 & 78.5 & 78.1 & 82.8 & 83.1 & 84.6 & 85.2 & 86.4 & 86.4 & 88.2 & 86.4 & \textbf{89.7} & 88.9 & \underline{89.0}  \\
        \noalign{\vskip 0.18ex}
        \noalign{\global\setlength{\arrayrulewidth}{0.8pt}}
        \cline{0-15}
        \noalign{\global\setlength{\arrayrulewidth}{0.4pt}}
    \end{tabular}
    \caption{Accuracy (\%) comparison on the ModelNet40 and ScanObjectNN datasets. Results marked with * represent our reproduced implementations using official codebases. \textbf{Bold} and \underline{underline} indicate the best and second-best results, respectively.}
    \label{tab1}
\end{table*}

\begin{table*}[ht]
  \centering
  \normalsize
  \setlength{\tabcolsep}{5pt} 
  \renewcommand{\arraystretch}{1.08}  
  \begin{tabular}{l| c c c c c | c c c c c}
    \noalign{\global\setlength{\arrayrulewidth}{0.8pt}}
    \cline{0-10}
    \rule{-2.5pt}{2.5ex}
    \multirow{2}{*}[-3pt]{\textbf{Method}} & \multicolumn{5}{c|}{\bfseries{ModelNet40} } & \multicolumn{5}{c}{\bfseries{ScanObjectNN} } \\
    \noalign{\global\setlength{\arrayrulewidth}{0.5pt}}
    \cline{2-11}
    \rule{0pt}{2.5ex}
     & 1-shot & 2-shot & 4-shot & 8-shot & 16-shot & 1-shot & 2-shot & 4-shot & 8-shot & 16-shot \\
    \noalign{\global\setlength{\arrayrulewidth}{0.5pt}}
    \cline{0-10}
    \rule{-2.5pt}{2.5ex}
    PointNet \cite{qi2017pointnet} &  27.4 & 32.3 & 55.1 & 64.8 & 72.3 & 18.8 & 26.2 & 26.6 & 35.0 & 35.2\\
    SimpleView \cite{goyal2021revisiting} & 27.5 & 36.1 & 58.8 & 68.3 & 78.4 & 23.0 & 23.6 & 29.6 & 32.0 & 37.4 \\
    CurveNet \cite{9711196} & 40.0 & 55.5 & 70.0 & 75.0 & 80.4 & 25.4 & 26.4 & 26.6 & 30.0 & 35.2 \\
    PointNet++ \cite{qi2017pointnet++} & 40.1 & 55.5 & 72.0 & 78.2 & 80.2 & 27.1 & 32.3 & 41.0 & 47.5 & 54.8 \\
    PointCLIP \cite{zhang2022pointclip} & 52.1 & 67.5 & 75.6 & 80.5 & 85.4 & 30.0 & 42.1 & 46.7 & 50.0 & 54.9 \\
    PointCLIP V2 \cite{zhu2023pointclip} & \underline{60.5} & 71.2 & 76.9 & 80.5 & 85.4 & 34.0 & 43.2 & 49.1 & 52.2 & 54.9 \\
    PPT \cite{sun204parameter} & 59.9 & \underline{73.8} & \underline{81.0} & \underline{86.1} & \underline{89.1} & \underline{35.2} & \underline{49.4} & \underline{57.7} & \underline{65.2} & \underline{73.9} \\
    \noalign{\vskip 0.18ex}
    \noalign{\global\setlength{\arrayrulewidth}{0.5pt}}
    \cline{0-10}
    \rule{-2.5pt}{2.5ex}
    \textbf{PCQ (Ours)} & \textbf{61.1} & \textbf{76.3} & \textbf{81.5} & \textbf{87.5} & \textbf{90.8} & \textbf{41.3} & \textbf{52.7} & \textbf{59.7} & \textbf{71.0} & \textbf{76.5} \\
    $\mathrm{\Delta}$  & +0.6 & +2.5 & +0.5 &+1.4 & +1.7 & +6.1 & +3.3 & +2.0 & +5.8 & +2.6\\
    \noalign{\global\setlength{\arrayrulewidth}{0.8pt}}
    \cline{0-10}
    \noalign{\global\setlength{\arrayrulewidth}{0.4pt}}
  \end{tabular}
  \caption{Few-shot accuracy (\%) comparison on ModelNet40 and ScanObjectNN (PB). 
 \textbf{Bold} and \underline{underline} indicate the best and second-best results, respectively.
  $\mathrm{\Delta}$ indicates the absolute improvement of our method over the second-best result.}
  \label{tab2}
\end{table*}

\begin{table}[t]
  \centering
  \normalsize
  \renewcommand{\arraystretch}{1.08}
  \setlength{\tabcolsep}{7pt}
  \begin{tabular}{lccc}
    \toprule
    \textbf{Method} & \textbf{10\%} & \textbf{20\%} & \textbf{100\%} \\
    \midrule
    PointNet \cite{qi2017pointnet}  & 72.7 & 73.5 & 80.4\\
    PointNet++ \cite{qi2017pointnet++} & 74.8 & 76.8 & 81.9\\
    PointCNN \cite{li2018pointcnn}   & 60.4 & 64.1 & 84.6\\
    PointBERT \cite{yu2022point}  & 76.4 & 79.6 & 84.1\\
    PPT \cite{sun204parameter}    & 80.8 & 84.0 & 86.4\\
    PCQ (Ours) & \textbf{82.6} & \textbf{84.9} & \textbf{86.6} \\
    \bottomrule
  \end{tabular}
  \caption{Mean class-wise IoU ($\mathbf{mIoU}_\mathcal{C}$) for part segmentation on ShapeNetPart under different training data ratios.}
  \label{tab:shapenetpart}
\end{table}

\subsection{Objective Function}
While multimodal alignment exhibits strong discriminative capabilities, it often suffers from high intra-class variance and insufficient inter-class separation in the embedding space.
We introduce a dual regularization strategy that simultaneously enforces prototype-wise compactness and separation.
As illustrated in Figure \ref{framework}, our framework employs a three-component constraint combining contrastive, compactness, and separation losses.

\noindent
\textbf{Contrastive Loss}. This forms the cornerstone of adaptive prompt tuning, designed to align the hybrid feature $\textbf{f}_i$ with its corresponding text feature $\textbf{h}^T_{y_i}$ in a shared embedding space.
Formally, the contrastive loss is defined as,
\begin{equation}
\label{loss_con}
\mathcal{L}_{\text{Align}} = -\frac{1}{N}\sum_{i=1}^N \log \frac{\exp\left (\cos ({\textbf{f}}_{i}, \textbf{h}^T_{y_i})\right)}{\sum_{j=1}^K \exp\left(\cos({\textbf{f}}_{i}, \textbf{h}^T_{j})\right)}.
\end{equation}

\noindent
\textbf{Compactness Loss}.
This loss aims to minimize intra-class variance by encouraging point cloud features to cluster tightly around assigned prototypes. The loss is defined as, 
\begin{equation}
\label{loss_intra}
\mathcal{L}_{\text{Comp}} = \| \textbf{H}^P - \textbf{Q} \textbf{H}^T \|^{2},
\end{equation}
where $\textbf{H}^P=[\textbf{h}_1^P;\textbf{h}_2^P;\ldots;\textbf{h}_N^P]\in \mathbb{R}^{N\times d}$ and 
$\textbf{H}^T=[\textbf{h}_1^T;\textbf{h}_2^T;\ldots;\textbf{h}_K^T]\in \mathbb{R}^{K\times d}$ denotes the matrix of point cloud features and textual prototype embeddings.
$\textbf{Q}=[\textbf{q}_1;\textbf{q}_2;\ldots;\textbf{q}_N]\in \mathbb{R}^{N\times K}$ is the assignment matrix where $\textbf{q}_i$ is a one-hot vector with $q_{ik}=1$ if $k=\arg\max_{k}y_{ik}$, i.e., the prototype with the highest similarity to the $i$-th sample, and 0 otherwise.
The compactness loss encourages point cloud features to align closely with their assigned text-guided prototypes, thereby reducing intra-class variance.

\newcolumntype{x}[1]{>{\centering\arraybackslash}p{#1pt}}
\newcommand{\tablestyle}[2]{\setlength{\tabcolsep}{#1}\renewcommand{\arraystretch}{#2}}
\begin{table*}[t]
\centering
\small
\tablestyle{2pt}{1.1}
\subfloat[Ablation on loss function components. \label{tab:ablations:loss}]{
\begin{tabular}{ccc@{\hspace{2mm}}c}
\toprule
$\mathcal{L}_{\text{A}}$ & $\mathcal{L}_{\text{S}}$ & $\mathcal{L}_{\text{C}}$ & Acc (\%) \\
\midrule
\faCheck &         &         & 69.95 \\
\faCheck & \faCheck &         & 69.19 \\
\faCheck &         & \faCheck & 70.01 \\
\faCheck & \faCheck & \faCheck & \textbf{71.03} \\
\bottomrule
\end{tabular}
}
\hspace{2mm}
\subfloat[Ablation on framework components. \textit{w/o} stands for without. \label{tab:ablations:components}]{
\begin{tabular}{p{2.9cm}c}
\toprule
\textbf{Network Component} & Acc (\%) \\
\midrule
\textit{w/o} PC adapter & 56.73 \\
\textit{w/o} Learnable prompt     & 67.66 \\
\textit{w/o} PC quantization  & 67.59 \\
Ours                    & \textbf{71.03} \\
\bottomrule
\end{tabular}
}
\hspace{2mm}
\subfloat[Comparison of prompt design strategies. \{class\} indicates the class name.\label{tab:ablations:prompts}]{
\begin{tabular}{p{3.7cm}c}
\toprule
\textbf{Prompt Design} & Acc (\%) \\
\midrule
\rule{0pt}{9.9pt}\parbox[c][14pt][c]{3.6cm}{\{class\}} & 67.66 \\
\rule{0pt}{9.9pt}\parbox[c][14pt][c]{3.6cm}{``A 3D shape of a '' + \{class\}} & 69.19 \\
\rule{0pt}{9.9pt}\parbox[c][14pt][c]{3.6cm}{Learnable prompt + \{class\}} & \textbf{71.03} \\
\bottomrule
\end{tabular}
}
\hspace{2mm}
\subfloat[Comparison of different prototype strategies.\label{tab:ablations:prototype}]{
\begin{tabular}{p{2.7cm}c}
\toprule
\textbf{Prototype Strategy} & Acc (\%) \\
\midrule
\rule{0pt}{9.9pt}\parbox[c][14pt][c]{2.7cm}{Cluster centroids} & 69.60 \\
\rule{0pt}{9.9pt}\parbox[c][14pt][c]{2.7cm}{Trainable codebooks} & 70.06 \\
\rule{0pt}{9.9pt}\parbox[c][14pt][c]{2.7cm}{Text embeddings}     & \textbf{71.03} \\
\bottomrule
\end{tabular}
}

\caption{Ablation study of our model on the ScanObjectNN-PB 8-shot setting. \textbf{Bold} indicates the best performance.}
\label{tab:ablations}
\end{table*}

\begin{table}[t]
  \centering
  \normalsize
  \renewcommand{\arraystretch}{1.08} 
  \setlength{\tabcolsep}{16pt}
  \begin{tabular}{l c c c}
    \toprule
    \textbf{Method} & BG & PB & MN40 \\
    \midrule
    PPT  & 83.2 & 70.6 & 18.8 \\
    \textbf{PCQ} & \textbf{86.9}& \textbf{72.8} & \textbf{21.5} \\
    $\mathrm{\Delta}$ & +3.7 & {+2.2} & {+2.7}  \\
    \bottomrule
  \end{tabular}
  \caption{ 
  Cross-dataset generalization with models trained on OBJ and evaluated on BG, PB, and ModelNet40.}
  \label{zero-shot}
\end{table}

\noindent
\textbf{Separation Loss}.
To maximize prototype discriminability, we enforce uniform geometric spacing among class prototypes in the embedding space.
This distributional regularization encourages prototypes to spread out evenly on the hypersphere, reducing inter-class confusion and improving generalization to unseen samples.
The constraint is formulated as minimizing the Kullback-Leibler divergence,
\begin{equation}
\mathcal{D}_{\mathrm{KL}}\left(P_{\text{proto}} \parallel \mathcal{U}\right) = \frac{1}{K}\sum_{k=1}^{K}\sum_{j \neq k} p_{kj}\log\frac{p_{kj}}{u_{kj}},
\end{equation}
where $P_{\text{proto}}$ denotes the probability distribution over prototypes and $ \mathcal{U}$ is the uniform distribution.
Each element is defined as,
\begin{equation}
p_{ij} = \frac{\exp\left(-\|\mathbf{h}_i^T - \mathbf{h}_j^T\|^2\right)}{\sum_{k \neq i} \exp\left(-\|\mathbf{h}_i^T - \mathbf{h}_k^T\|^2\right)}.
\end{equation}
Minimizing \(\mathcal{D}_{\mathrm{KL}}\) drives the learned prototype distribution \(p_{ij}\) toward uniformity, encouraging approximately equal pairwise distances. When pairwise distances are balanced, the total inter-prototype dispersion is maximized, leading to enhanced class separability. We thus derive the separation loss as,
\begin{equation}
\mathcal{L}_{\mathrm{Sep}} = \sum_{i \neq j} \exp\left(-\|\mathbf{h}_i^T - \mathbf{h}_j^T\|^2\right).
\end{equation}

We introduce hyperparameters $\lambda_1$ and $\lambda_2$ to balance three complementary components of our objective function. The overall loss is  formulated as,
\begin{equation}
\mathcal{L}_{\text{Total}} = \mathcal{L}_{\text{Align}} + \lambda_1 \mathcal{L}_{\text{Comp}} + \lambda_2 \mathcal{L}_{\text{Sep}}.
\end{equation}
The multi-loss framework has three objectives: the contrastive loss reduces intra-class variance by pulling together positive pairs while pushing apart negative samples across classes. The compactness loss enforces tight clustering of point cloud features, strengthening local feature aggregation, and the separation loss maximizes prototype distances. 

\section{Experiments}

\subsection{Main Results}
\textbf{Point Cloud Recognition.} 
As shown in Table~\ref{tab1}, our approach achieves strong performance across all benchmark datasets while maintaining high parameter efficiency.
Despite using fewer trainable parameters under the Parameter-Efficient Fine-Tuning (PEFT) paradigm, our method delivers competitive results compared to both fully supervised and fully fine-tuned approaches.
Notably, on the OBJ variant, our method surpasses the current state-of-the-art PEFT approach by +0.4\%.
For the PB variant, while our accuracy is lower than that of fully fine-tuned ULIP-2, it uses only 8.0\% of ULIP-2's trainable parameters, highlighting a good trade-off between performance and model complexity.

\noindent
\textbf{Few-Shot Recognition.}
Following prior protocols \cite{zhu2023pointclip}, we evaluate our method under 1-shot to 16-shot settings, sampling 1–16 instances per class for training and testing on the full test set.
As listed in Table~\ref{tab2}, our approach achieves state-of-the-art performance across all settings and exhibits strong gains in extreme data scarcity, e.g., +6.1\% improvement on 1-shot ScanObjectNN.  
The performance advantage is most pronounced in low-shot scenarios, indicating strong generalization and sample efficiency under limited supervision.
Moreover, consistent improvements on both synthetic and real-world datasets demonstrate the robustness and broad applicability of our framework.

\noindent
\textbf{Shape Part Segmentation.}
We evaluate 3D part segmentation performance on the ShapeNetPart dataset under varying levels of supervision.
To adapt our framework for segmentation, we attach a part segmentation head to the 3D encoder.
The model is trained using different proportions of the labeled training data, and we report the mean class-wise IoU ($\mathbf{mIoU}_\mathcal{C}$) as the evaluation metric.
As shown in Table~\ref{tab:shapenetpart}, our method achieves the highest performance across all supervision settings, attaining $\mathbf{mIoU}_\mathcal{C}$ scores of 82.6\%, 84.9\%, and 86.6\% when trained with 10\%, 20\%, and 100\% of the data, respectively.
These results demonstrate not only the strong segmentation capability of our learned representations but also their effectiveness in low-data regimes.

\begin{table}[t]
  \centering
  \normalsize
  \setlength{\tabcolsep}{4 pt} 
  \renewcommand{\arraystretch}{1.1}  
  \begin{tabular}{l c c c c c }
    \toprule
    \textbf{Method} & 1-shot & 2-shot & 4-shot & 8-shot & 16-shot \\
    \midrule
    Uni3D (PEFT) & 45.07 & 53.68 & 56.49 & 65.58 & 68.56 \\
    \textbf{Uni3D (PCQ)} & \textbf{50.17} & \textbf{56.42} & \textbf{57.11} & \textbf{65.93} & \textbf{71.20} \\
    $\mathrm{\Delta}$ & +5.10 & +2.74 & +0.62 & +0.35 & +2.64\\
    \bottomrule
  \end{tabular}
  \caption{Few-shot accuracy (\%) of PEFT and PCQ on Uni3D backbone. $\mathrm{\Delta}$ indicates the absolute accuracy improvement.}
  \label{tab_uni3d}
\end{table}

\subsection{Ablation Studies}
We conduct ablations on ScanObjectNN-PB with an 8-shot setting to balance data scarcity and evaluation stability.

\noindent\textbf{Loss Function Analysis.} 
As listed in Table~\ref{tab:ablations:loss}, we evaluate the contribution of each loss component. Using only the alignment loss $\mathcal{L}_{\text{Align}}$ as the baseline achieves 69.95\% accuracy. Incorporating the compactness loss $\mathcal{L}_{\text{Comp}}$ brings a marginal improvement of 0.06\%, suggesting that optimizing intra-class compactness alone has a limited effect on overall discriminability. In contrast, adding the separation loss $\mathcal{L}_{\text{Sep}}$ leads to a 0.76\% decrease, indicating that optimizing inter-class separation in isolation can disrupt intra-class coherence and harm generalization.
When all three losses are combined, the model achieves an accuracy of 71.03\%, highlighting the importance of dual regularization. 

\noindent\textbf{Component Analysis.} 
As shown in Table \ref{tab:ablations:components}, the learnable text prompts enable adaptive refinement of textual representations, while the point cloud adapter fine-tunes the visual encoder to improve cross-modal alignment. 
The point cloud quantization module discretizes continuous visual features into text-guided prototypes, and we integrate these features with textual prototypes.
The removal of any component results in a significant performance drop.

\noindent\textbf{Textual Prompt Analysis.} 
Table \ref{tab:ablations:prompts} presents an evaluation of different textual prompt designs, including the bare class name \{class\}, the handcrafted template ``A 3D shape of a '' + \{class\}, and Learnable prompts + \{class\}.
The fixed template yields a 1.53\% accuracy improvement over the baseline, demonstrating the necessity of structural prompts. Notably, learnable prompts achieve a significant performance gain of 1.84\% over fixed templates, showing that prompts optimized via backpropagation possess stronger task adaptability and cross-modal alignment capabilities.

\noindent\textbf{Prototype Strategy Analysis.}  
As shown in Table \ref{tab:ablations:prototype}, we evaluate three prototype learning strategies: cluster centroids, trainable codebooks, and text embeddings. 
Using cluster centroids introduces dependency on the training data, leading to a 1.43\% accuracy drop under distribution shifts.
Trainable codebooks optimize prototypes through gradient descent but show unstable convergence, achieving only 70.06\% accuracy.
In contrast, our approach addresses these limitations, reaching 71.03\% accuracy with the large-scale pretraining for multimodal alignment.

\begin{figure}[t]
  \centering
    \centering
    \includegraphics[width=0.94\linewidth]{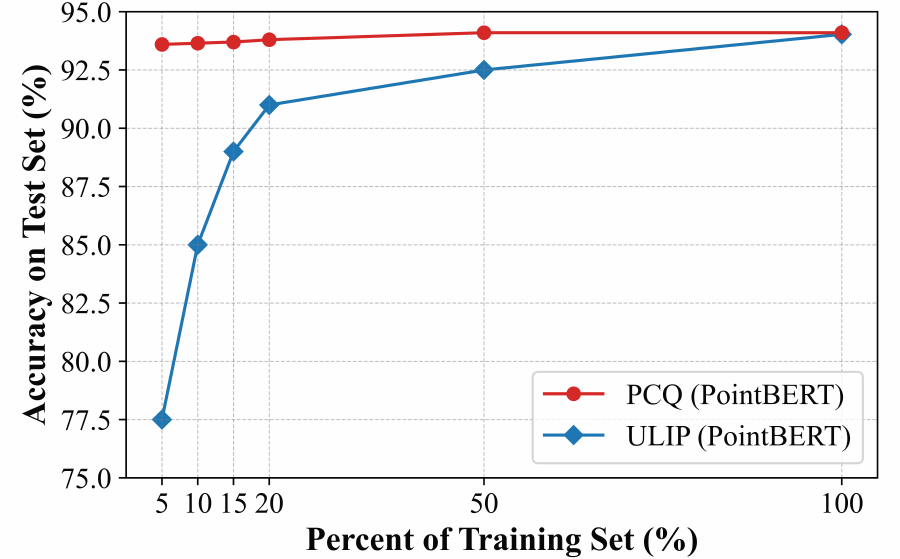}
    \caption{Data efficiency comparison. Models are trained on varying percentages of data and evaluated on the full test set.}
    \label{data_efficiency}
    \bigskip
    \centering
    \includegraphics[width=\linewidth]{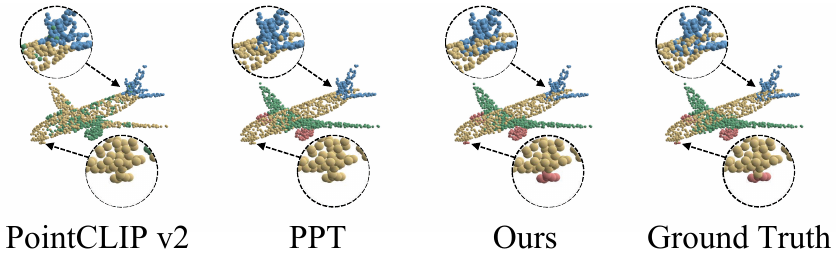}
    \caption{Part segmentation visualization on ShapeNetPart.}
    \label{vis_partseg}
\end{figure}

\subsection{More Analyses}

\noindent\textbf{Data Efficiency Analysis.}
Large-scale pre-trained models show the potential to reduce dependence on labeled data for downstream adaptation. We evaluate the data efficiency of our parameter-efficient tuning framework through comparisons with the ULIP baseline. 
We conduct experiments on ModelNet40 by progressively enlarging training subsets (5\%, 10\%, 15\%, 20\%, 50\%, 100\%) while keeping the full test set fixed for all experiments. As illustrated in Figure \ref{data_efficiency}, our approach outperforms the baseline in low-data scenarios, achieving 93.6\% accuracy with merely 5\% training samples.

\noindent\textbf{Cross-Dataset Generalization Analysis.}
As shown in Table~\ref{zero-shot}, we train our model on the OBJ variant of ScanObjectNN, and evaluate its performance on the BG and PB variants and the ModelNet40 dataset. Our method consistently outperforms the baseline, with accuracy gains of +3.7\% on BG, +2.2\% on PB, and +2.7\% on ModelNet40.
These results prove that our approach is robust across dataset variants and generalizes well to out-of-domain data.

\noindent\textbf{Backbone Analysis.}
To assess the generality of our approach across different architectures, we integrate PCQ into the Uni3D‑Ti \cite{zhou2023uni3d} backbone and evaluate it on ScanObjectNN (PB) under few-shot settings. As shown in Table \ref{tab_uni3d}, PCQ consistently outperforms standard PEFT with adapters across all shots, proving its architecture-agnostic nature and ability to deliver stable performance gains.

\noindent\textbf{Part Segmentation Visualizations.}
As depicted in Figure \ref{vis_partseg}, our method surpasses both PointCLIP v2 and PPT in ShapeNetPart part segmentation, with particularly notable improvements in fine-grained regions such as the tail and landing gear. The predicted boundaries are sharper, evidencing a superior ability to capture fine-grained details. 

\begin{figure}[t] 
\centering 
\includegraphics[width=1.0\columnwidth]{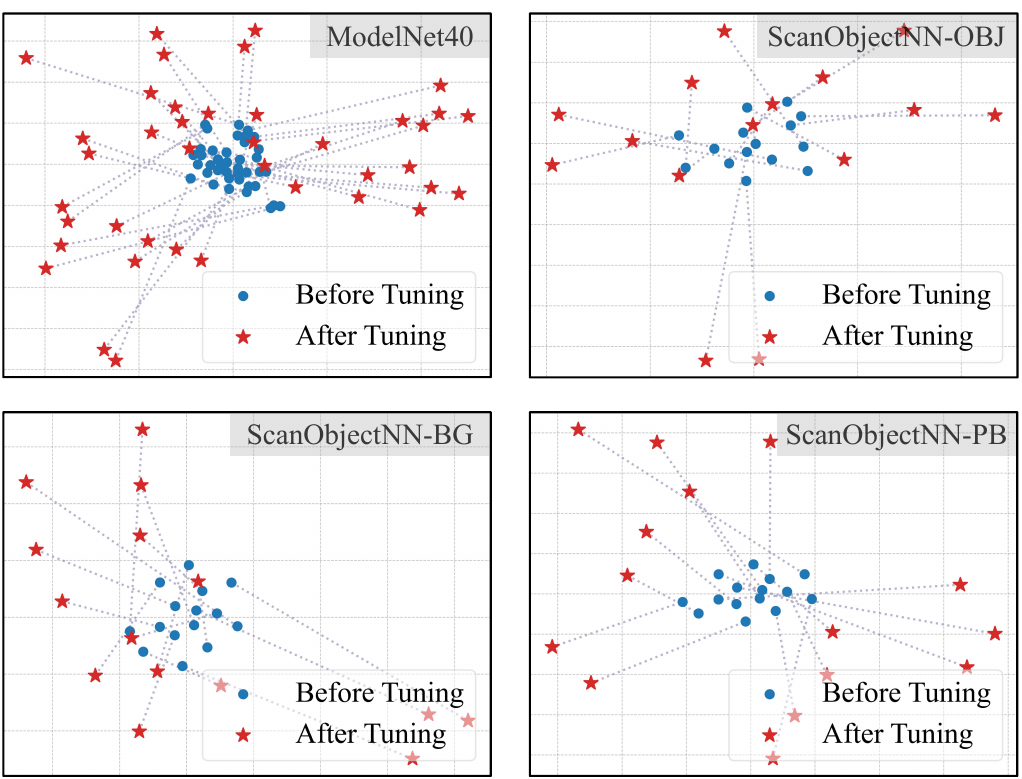} 
\caption{The t-SNE visualization before (\raisebox{-0.2ex}{\includegraphics[height=1.5ex]{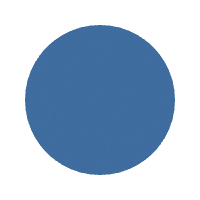}}) and after (\raisebox{-0.6ex}{\includegraphics[height=2.5ex]{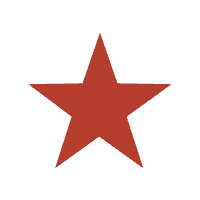}}) model fine-tuning across four datasets. Dashed lines connect corresponding prototypes across training phases.} 
\label{visualization} 
\end{figure}

\noindent\textbf{Prototype Visualization.} 
We use t-SNE to visualize prototype embeddings. 
Connection lines indicate prototype evolution trajectories, and distances between prototypes reflect the degree of class separation.
As shown in Figure \ref{visualization}, prototypes are tightly clustered and poorly separated before fine-tuning.
After fine-tuning, the prototypes exhibit improved inter-class separation, forming well-separated clusters. This increased separability reduces quantization mismatches.

\section{Conclusion}

In this paper, we propose a text-driven point cloud quantization framework that enhances 3D understanding by leveraging vision-language alignment.
We introduce the Gumbel-Softmax relaxation to discretize continuous point cloud features into learnable textual prototypes, enabling differentiable quantization while promoting semantic interpretability. A hybrid mechanism is then designed to effectively integrate quantized prototypes with original features. We further introduce dual regularizations, in which the compactness loss minimizes intra-class variance, while the separation loss reduces inter-class overlaps.
Experiments on benchmarks validate the effectiveness of the proposed approach. In the future, we plan to explore dynamic prototype generation for fine-grained part-level correspondence.

\section{Acknowledgments}
This work was supported by the National Key Research and Development Program of China under Grant 2025YFA0921700, the National Natural Science Foundation of China (No.s 62406221, 62222608, 62436002), the Tianjin Natural Science Funds for Distinguished YoungScholar (No.23JCIQIC00270), the Natural Science Foundation of Tianjin (No.25JCQNJC00770), and the 2024 Open Research Project of Intelligent Policing and National Security Risk Management Laboratory under Grant ZHKFYB2404.

\bibliography{aaai2026}
\appendix
\input{appendix}

\end{document}

%% file: appendix.tex
\appendix

\section{Training Algorithm}\label{app:pseudocode}
Algorithm~\ref{alg:pcq} summarizes the training procedure of the Point Cloud Quantization (PCQ) framework. The definitions of all loss terms are provided in Eqs.~(8)--(10) in the main text.

\begin{algorithm}[h]
\footnotesize
\caption{Training Procedure of PCQ}
\label{alg:pcq}
\DontPrintSemicolon
\SetAlgoLined
\KwIn{
Class names $\mathcal{C}=\{\textbf{c}_k\}_{k=1}^{K}$,
Point cloud $\mathcal{P} = \{ P_i \}_{i=1}^{N}$,
Pre-trained text encoder $\mathcal{F_T}(\cdot)$, 
Pre-trained point cloud encoder $\mathcal{F_P}(\cdot)$.
}
\vspace{0.3em}
\KwOut{
Fine-tuned point cloud encoder $\mathcal{F_{P'}}(\cdot)$, 

\hspace*{3.7em} Learned prompts vectors $\textbf{u}= [\textbf{u}_1, \textbf{u}_2, \ldots , \textbf{u}_m]$.
}
\vspace{0.3em}
\For{$\text{epoch}=1$ \KwTo $\text{Epoch}$}{
  \ForEach{mini-batch $\mathcal{B}=\{(P_i,y_i)\}_{i=1}^{B}$}{
    \For{$k=1$ \KwTo $K$}{
        \emph{\% build textual prototypes}   
        
        \hspace*{0.5em}$\textbf{h}^T_k = \mathcal{F_T}([\textbf{u}_1,\textbf{u}_2,\dots,\textbf{u}_m,\,\textbf{c}_k])$
    }
  \For{$i=1$ \KwTo $B$}{
    \emph{\% extract point cloud features}   

    \hspace*{0.5em}$\textbf{h}^P_i = \mathcal{F}_{P'}(P_i)$

    \vspace{0.3em}

    \emph{\% differentiable quantization}  
    
    \hspace*{0.5em}$\mathbf{s}_i = \cos(\textbf{h}^P_i, \textbf{h}^T_{1:K})$
    
    \hspace*{0.5em}$[y_{i1}, ..., y_{iK}] = \text{GumbelSoftmax}(\mathbf{s}_i, \tau)$
    
    \hspace*{0.5em}$\textbf{v}_i = \sum_{k=1}^K y_{ik} \textbf{h}^T_k$

}

    \emph{\% cross-modal feature fusion}   

    \hspace*{0.5em}$\textbf{f}_i = \text{FFN}(\text{CrossAttention}(\textbf{h}^P_i,\textbf{v}_i))+\textbf{h}^P_i$
    
    \vspace{0.3em}
    \emph{\% compute losses}  

    \hspace*{0.5em}$\mathcal{L}_{\text{Align}}\gets$  Contrastive loss between $\textbf{f}$ and $
    \textbf{h}^{T}$ as Eq. (8)

    \hspace*{0.5em}$\mathcal{L}_{\text{Comp}} \gets$ Intra-class compactness loss as Eq. (9)

    \hspace*{0.5em}$\mathcal{L}_{\text{Sep}} \gets$ Inter-class separability loss as Eq. (10)

    \hspace*{0.5em}$\mathcal{L}_{\text{Total}} =
      \mathcal{L}_{\text{Align}} +
      \lambda_1\mathcal{L}_{\text{Comp}} +
      \lambda_2\mathcal{L}_{\text{Sep}}$
      
    \vspace{0.3em}  
    \emph{\% update trainable params}  
    
    \hspace*{0.5em}Update $\theta_{\mathcal{P'}}$ and $\mathbf{u}_{1:m}$ using gradients of $\mathcal{L}_{\text{Total}}$

  }
}
\end{algorithm}

\section{Experimental Setup}\label{app:setup}

\subsection{Datasets}
We conducted extensive experiments on two widely adopted 3D point cloud datasets, including ModelNet40 \footnote{Wu et al. 2015. 3d shapenets: A Deep Representation for Volumetric Shapes. In \textit{CVPR}, 1912–1920.} and ScanObjectNN \footnote{Uy et al. 2019. Revisiting Point Cloud Classification: A New Benchmark Dataset and Classification Model on Real-World Data. In \textit{ICCV}, 1588–1597.}.

\noindent\textbf{ModelNet40} is a standard benchmark for point cloud analysis, comprising 12,311 high-quality synthetic CAD models from 40 object categories. The dataset is split into 9,843 training and 2,468 test samples, with point clouds uniformly sampled from mesh surfaces to ensure geometric coverage.

\noindent\textbf{ScanObjectNN} provides a more challenging and real-world benchmark for point cloud analysis, containing 2,902 objects across 15 categories. The dataset captures inherent challenges such as occlusions and background noise. For fair evaluation, we assessed on three variants: OBJ ONLY (OBJ) contains isolated objects; OBJ BG (BG) includes objects with background; and PB T50 RS (PB) with random perturbations and clutter, making it the most challenging variant. 

\subsection{Implementation Details}
All experiments were conducted on two NVIDIA RTX 3090 GPUs. The models were trained for 250 epochs with a batch size of 30, using the AdamW optimizer with an initial learning rate of 0.003. 
The learning rate was linearly warmed up over the first 10 epochs and then decayed via cosine annealing.
Each point cloud sample comprised 1024 points, processed using PointBERT as the backbone for the point cloud encoder. 
During training, the parameters of both the visual and text encoders were frozen,  except for the text prompts and the last Transformer block of the point cloud encoder. The length of the text prompt was set to 32, and both hyperparameters $\lambda_1$ and $\lambda_2$ were set to 0.01.

\begin{figure}[t] 
\centering 
\includegraphics[width=0.95\linewidth]{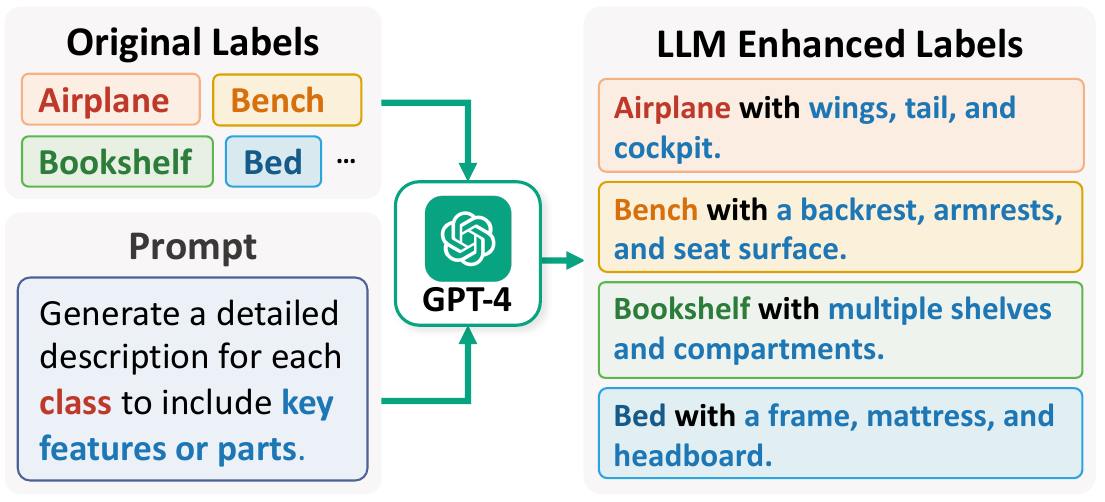} 
\caption{Label augmentation by employing the large language model GPT-4 to provide enriched textual descriptions. 
} 
\label{enhance_label} 
\end{figure}

\begin{table}[t]
  \centering
  \renewcommand{\arraystretch}{1}
  \begin{tabular}{l c c c}
    \toprule
    Setting & Class Label & Label Enrichment & $\mathrm{\Delta}$ \\
    \midrule
    2-shot   & 52.74 & 55.52 & \textbf{+2.78} \\
    8-shot   & 71.03 & 72.07 & \textbf{+1.04} \\
    32-shot  & 81.54 & 81.96 & \textbf{+0.42} \\
    128-shot & 86.33 & 86.57 & \textbf{+0.24} \\
    \bottomrule
  \end{tabular}
  \caption{Ablation study on label enrichment.}
  \label{tab:enhance_label}
\end{table}

\section{Additional Experiments}
\subsection{Label Enrichment Analysis}
As illustrated in Figure \ref{enhance_label}, we leveraged GPT-4 to augment class labels with semantically relevant attributes, generating enhanced textual descriptions that more comprehensively capture object characteristics.
Table \ref{tab:enhance_label} shows that label enrichment brings substantial performance gains in extremely low-data settings, confirming the benefits of supplementing limited visual input with complementary linguistic priors.
As the number of training samples increases, however, the performance gap between 
base and enhanced labels gradually narrow.
With sufficient point cloud data, our framework achieves effective concept learning through cross-modal alignment, reducing the reliance on external textual knowledge.
To balance efficiency and simplicity without sacrificing generalization, we adopt base class labels.

\begin{figure*}[ht] 
\centering 
\includegraphics[width=0.8\linewidth]{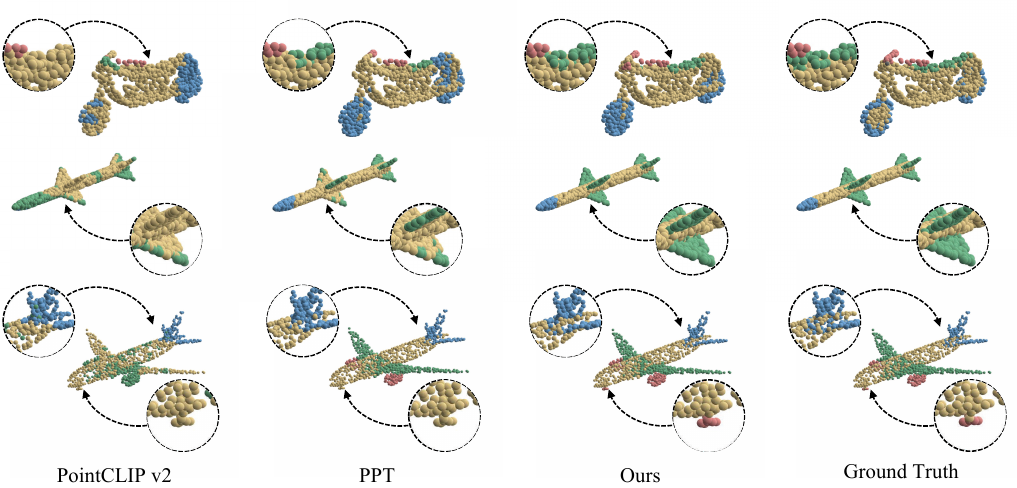} 
\caption{Additional part segmentation visualizations. Zoomed-in regions highlight key differences. 
} 
\label{fig:appendix_vis} 
\end{figure*}

\subsection{Gumbel-Softmax Temperature Analysis}
We conducted an ablation study on the temperature parameter $\tau$ in Gumbel-Softmax relaxation to investigate its impact on model performance. A small $\tau$ 
encourages sharp, near one-hot distributions, which leads to hard assignments.
While this enhances decision definiteness, it may hinder gradient flow and reduce the model’s ability to correct early or uncertain predictions. 
In contrast, a large $\tau$ results in overly smooth distributions, making it harder to distinguish between classes. 
An intermediate value of $\tau$ strikes a balance between flexibility and discriminative power in the quantization process. Table~\ref{tab:temperature} indicates that setting $\tau=1$ achieves the best performance among the tested values.

\begin{table}[t]
    \centering
    \normalsize
    \renewcommand{\arraystretch}{1}
    \setlength{\tabcolsep}{4.5pt}
    \begin{tabular}{lcccccc}
        \toprule
        \textbf{$\tau$} & 0.1 & 0.3 & 0.5 & 1 & 3 & 5 \\
        \midrule
        Acc (\%) & 70.57 & 70.68 & 70.82 & \textbf{71.03} & 69.67 & 69.57 \\
        \bottomrule
    \end{tabular}
    \caption{Ablation study on the temperature parameter $\tau$ in Gumbel-Softmax.}
    \label{tab:temperature}
\end{table}

\subsection{Fine-Tuning Scope Analysis}
As shown in Table~\ref{tab:scope}, we conducted an ablation study on different fine-tuning scopes of the point cloud encoder. Fine-tuning only the final Transformer block achieves the highest accuracy, while tuning only the MLP layers yields inferior results, possibly due to their limited capacity for modeling global features. Increasing the number of fine-tuned Transformer blocks does not bring further gains. In contrast, fine-tuning all layers results in a significant drop in performance, indicating that extensive parameter updates can disrupt the geometric knowledge during large-scale pretraining.
\begin{table}[ht]
    \centering
    \renewcommand{\arraystretch}{1}
    \setlength{\tabcolsep}{3.3pt}
    \begin{tabular}{lcccccc}
        \toprule
        {\textbf{Setting}} & {\small 2MLP} & {\small 4MLP} & {\small 1Block} & {\small 2Block} & {\small 3Block} & {\small All layers} \\
        \midrule
        Acc(\%) & 64.19 & 69.71 & \textbf{71.03} & 70.52 & 70.57 & 20.78 \\
        \bottomrule
    \end{tabular}
    \caption{Ablations on fine-tuning scopes of the PC encoder. 2/4MLP denotes tuning the last 2 or 4 MLP layers; 1–3Block denotes tuning the last 1-3 Transformer blocks.}
    \label{tab:scope}
\end{table}

\subsection{Prototype Interpretability Analysis}
To quantitatively evaluate the interpretability of the learned prototypes, we introduce the \textit{Prototype Assignment Accuracy} (PAA), a metric that measures the consistency between assigned prototypes and ground-truth semantic categories among correctly classified samples. Formally,
\begin{equation}
\text{PAA} = \frac{1}{|\mathcal{C}|} \sum_{i \in \mathcal{C}} \mathbb{I}(p_i = y_i),
\end{equation}
where $\mathcal{C}$ denotes the set of correctly classified samples, $p_i$ is the prototype assigned to the $i$-th sample, $y_i$ is its ground-truth label, and $\mathbb{I}(\cdot)$ is the indicator function. As shown in Table \ref{tab:paa}, our model achieves a PAA of 93.8\%, indicating that the vast majority of prototype assignments align with the true semantic categories and thus demonstrate that the learned prototypes are not only functionally effective but also semantically meaningful.

\begin{table}[t]
    \centering
    \renewcommand{\arraystretch}{1}
    \setlength{\tabcolsep}{28pt}
    \begin{tabular}{lc}
        \toprule
        \textbf{Prototype Strategy} & PAA (\%) \\
        \midrule
        Cluster centroids &  90.4\\
        Trainable codebooks & 87.7 \\
        Text embeddings & \textbf{93.8} \\
        \bottomrule
    \end{tabular}
    \caption{Comparison of Prototype Assignment Accuracy (PAA) for different prototype learning strategies.}
    \label{tab:paa}
\end{table}

\section{Additional Visualizations}
Figure~\ref{fig:appendix_vis} shows additional part segmentation visualizations. Key regions are zoomed in to highlight differences. Our method provides more accurate and consistent segmentations, especially on fine-grained structures.